\documentclass[11pt,a4paper]{article}
\usepackage{eacl2017}
\usepackage{times}
\usepackage{amsmath,amssymb,amsfonts,amsthm,mathabx,latexsym}
\usepackage{multirow,array,booktabs,verbatim}
\usepackage{graphicx,xcolor,tikz,tikz-qtree,linguex,microtype}
\usepackage[inline]{enumitem}
\usepackage[hypcap]{caption}
\PassOptionsToPackage{hyphens}{url}
\usepackage[pdftex,unicode,hidelinks,breaklinks,hyperfootnotes=false,bookmarks]{hyperref}
\xdefinecolor{darkgreen}{RGB}{0, 128, 0}

\eaclfinalcopy
\def\eaclpaperid{114} %  Enter the ACL Paper ID here  %% FIXME

\title{A Data-Oriented Model of Literary Language}

\author{Andreas van Cranenburgh \\
  Institut f\"ur Sprache und Information \\
  Heinrich Heine University D\"usseldorf \\
  \texttt{cranenburgh@phil.hhu.de} \\\And
  Rens Bod \\
  Institute for Logic, Language and \\
  Computation, University of Amsterdam  \\
  \texttt{rens.bod@uva.nl} \\}
\date{}

\begin{document}

\maketitle

\begin{abstract}
We consider the task of predicting how literary a text is,
with a gold standard from human ratings.
Aside from a standard bigram baseline, we apply rich
syntactic tree fragments, mined from the training set,
and a series of hand-picked features.
Our model is the first to distinguish degrees of highly and less literary novels using
a variety of lexical and syntactic features,
and explains 76.0~\% of the variation in literary ratings.
\end{abstract}

\section{Introduction}

What makes a literary novel \emph{literary}?
This seems first of all to be a value judgment;
but to what extent is this judgment arbitrary,
determined by social factors,
or predictable as a function of the text?
The last explanation is associated with the concept of \emph{literariness},
the hypothesized linguistic and formal properties that distinguish literary
language from other language~\cite{baldick2008oxford}.
Although the definition and demarcation of literature
is fundamental to the field of literary studies,
it has received surprisingly little empirical study.
Common wisdom has it that literary distinction is attributed
in social communication about novels and that it lies mostly outside
of the text itself~\cite{bourdieu1996rules},
but an increasing number of studies argue that in addition to social
and historical explanations, textual features of various complexity
may also contribute to the perception of literature by readers
(cf.~Harris, 1995\nocite{harris1995literary}; McDonald, 2007\nocite{mcdonald2007critic}).
The current paper shows that not only lexical features but also hierarchical
syntactic features and other textual characteristics contribute to explaining
judgments of literature.

Our main goal in this project is to answer the following question:
are there particular textual conventions in literary novels
that contribute to readers judging them to be literary?
We address this question by building a model of literary evaluation
to estimate the contribution of textual factors.
This task has been considered before with a smaller set of novels
(restricted to thrillers and literary novels),
using bigrams~\cite{vancranenburgh2015identifying}.
We extend this work by testing on a larger, more diverse corpus, % of novels,
and by applying rich syntactic features and several hand-picked features
to the task.
This task is first of all relevant to literary studies---to reveal to what
extent literature is empirically associated with textual characteristics.
However, practical applications are also possible; e.g., an automated model
could help a literary publisher decide whether the work of a new author fits its audience;
or it could be used as part of a recommender system for readers.

Literary language is arguably a subjective notion.
A gold standard could be based on the expert opinions of critics
and literary prizes, but we can also consider the reader directly,
which, in the form of a crowdsourced survey, more easily provides
a statistically adequate number of responses.
We therefore base our gold standard on a large online survey of readers
with ratings of novels.

Literature comprises some of the most rich and sophisticated language,
yet stylometry typically does not exploit linguistic information
beyond part-of-speech (\textsc{pos}) tags or grammar productions,
when syntax is involved at all~(cf.\ e.g., Stamatatos et al., 2009\nocite{stamatatos2009survey}; Ashok et al., 2013\nocite{ashok2013success}).
While our results confirm that simple features are highly effective,
we also employ full syntactic analyses and argue for their usefulness.
We consider tree fragments:
arbitrarily-sized connected subgraphs of parse trees~\cite{swanson2012native,bergsma2012stylometric,vancranenburgh2012literary}.
Such features are central to the Data-Oriented Parsing
framework~\cite{scha1990,bod1992computational}, which postulates that
language use derives from arbitrary chunks (e.g., syntactic tree fragments)
of previous language experience.
In our case, this suggests the following hypothesis.

\textsc{Hypothesis 1:} Literary authors employ a distinctive inventory of
lexico-syntactic constructions (e.g., a register)
that marks literary language.

Next we provide an analysis of these constructions %this register of tree fragments
which supports our second hypothesis.

\textsc{Hypothesis 2:} Literary language invokes a larger set of syntactic constructions
when compared to the language of non-literary novels,
and therefore more variety is observed in the parse tree fragments
whose occurrence frequencies are correlated with literary ratings.

The support provided for these hypotheses suggests that the notion of
literature can be explained, to a substantial extent, from textual
factors, which contradicts the belief that external, social factors
are more dominant than internal, textual factors. 

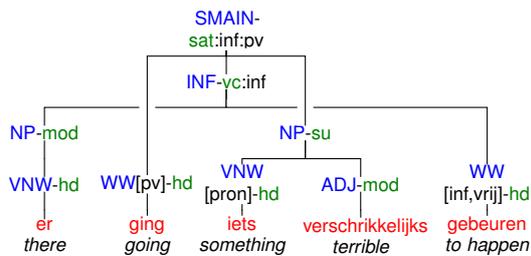
\begin{figure}\centering
	%\documentclass{article}\usepackage{tikz}\begin{document}

% discodop treetransforms --unbinarize --fmt=discbracket Franzen_Correcties.dbr | less
% (ROOT (DU^pv (SMAIN-nucl^inf^pv (NP-su (vnw-hd/VNW[pron] 0)) (ww-hd/WW[pv] 1) (INF-vc^inf (NP-obj1 (vnw-hd/VNW[pron] 2)) (ww-hd/WW[inf,vrij] 3))) (let/LET[] 4) (SMAIN-sat^inf^pv (INF-vc^inf (NP-mod (vnw-hd/VNW[] 5)) (ww-hd/WW[inf,vrij] 9)) (ww-hd/WW[pv] 6) (NP-su (vnw-hd/VNW[pron] 7) (adj-mod/ADJ[] 8)))) (let/LET[]^. 10))
% Je kon het voelen : er ging iets verschrikkelijks gebeuren .
\begin{tikzpicture}[yscale=0.85, xscale=1.3, scale=0.8, align=center,
		text width=1.5cm, inner sep=0mm, node distance=1mm]
%\footnotesize\sffamily
\scriptsize\sffamily
\path
	% (6, 6) node (n0) { \textcolor{blue}{ROOT} }
	% (5, 5) node (n1) { \textcolor{blue}{DU}:pv }
	% (1, 4) node (n2) { \textcolor{blue}{SMAIN}-\textcolor{darkgreen}{nucl}:inf:pv }
	% (0, 2) node (n3) { \textcolor{blue}{NP}-\textcolor{darkgreen}{su} }
	% (3.25, 3) node (n8) { \textcolor{blue}{INF}-\textcolor{darkgreen}{vc}:inf }
	% (2.5, 2) node (n9) { \textcolor{blue}{NP}-\textcolor{darkgreen}{obj1} }
	(8, 4) node (n16) { \textcolor{blue}{SMAIN}-\textcolor{darkgreen}{sat}:inf:pv }
	(8, 3) node (n17) { \textcolor{blue}{INF}-\textcolor{darkgreen}{vc}:inf }
	(9, 2) node (n25) { \textcolor{blue}{NP}-\textcolor{darkgreen}{su} }
	% (0, 1) node (n4) { \textcolor{blue}{VNW} [pron]-\textcolor{darkgreen}{hd} }
	% (0, 0) node (n5) { \textcolor{red}{Je} \\ \em You }
	% (1.25, 1) node (n6) { \textcolor{blue}{WW} [pv]-\textcolor{darkgreen}{hd} }
	% (1.25, 0) node (n7) { \textcolor{red}{kon} \\ \em could }
	% (2.5, 1) node (n10) { \textcolor{blue}{VNW} [pron]-\textcolor{darkgreen}{hd} }
	% (2.5, 0) node (n11) { \textcolor{red}{het} \\ \em it }
	% (4, 1) node (n12) { \textcolor{blue}{WW} [inf,vrij]-\textcolor{darkgreen}{hd} }
	% (4, 0) node (n13) { \textcolor{red}{voelen} \\ \em feel }
	% (5, 1) node (n14) { \textcolor{blue}{LET} }
	% (5, 0) node (n15) { \textcolor{red}{:} \\ \em : }
	(5.7, 2) node (n18) { \textcolor{blue}{NP}-\textcolor{darkgreen}{mod} }
	(5.7, 1) node (n19) { \textcolor{blue}{VNW}-\textcolor{darkgreen}{hd} }
	(5.7, 0) node (n20) { \textcolor{red}{er} \\ \em there }
	(7, 1) node (n23) { \textcolor{blue}{WW}[pv]-\textcolor{darkgreen}{hd} }
	(7, 0) node (n24) { \textcolor{red}{ging} \\ \em going }
	(8.2, 1) node (n26) { \textcolor{blue}{VNW} [pron]-\textcolor{darkgreen}{hd} }
	(8.2, 0) node (n27) { \textcolor{red}{iets} \\ \em something }
	(9.7, 1) node (n28) { \textcolor{blue}{ADJ}-\textcolor{darkgreen}{mod} }
	(9.7, 0) node (n29) { \textcolor{red}{verschrikkelijks} \\ \em terrible }
	(11.3, 1) node (n21) { \textcolor{blue}{WW} [inf,vrij]-\textcolor{darkgreen}{hd} }
	(11.3, 0) node (n22) { \textcolor{red}{gebeuren} \\ \em to happen}
	% (12.95, 1) node (n30) { \textcolor{blue}{LET}:. }
	% (12.95, 0) node (n31) { \textcolor{red}{.} \\ \em .}
;
% \draw (n30) -- +(0, -0.5) -| (n31);
\draw (n28) -- +(0, -0.5) -| (n29);
\draw (n26) -- +(0, -0.5) -| (n27);
\draw (n23) -- +(0, -0.5) -| (n24);
\draw (n21) -- +(0, -0.5) -| (n22);
\draw (n19) -- +(0, -0.5) -| (n20);
% \draw (n14) -- +(0, -0.5) -| (n15);
% \draw (n12) -- +(0, -0.5) -| (n13);
% \draw (n10) -- +(0, -0.5) -| (n11);
% \draw (n6) -- +(0, -0.5) -| (n7);
% \draw (n4) -- +(0, -0.5) -| (n5);
\draw (n25) -- +(0, -0.5) -| (n26);
\draw (n25) -- +(0, -0.5) -| (n28);
\draw (n18) -- +(0, -0.5) -| (n19);
% \draw (n9) -- +(0, -0.5) -| (n10);
% \draw (n3) -- +(0, -0.5) -| (n4);
\draw [white, -, line width=6pt] (n17)  +(0, -0.5) -| (n18);
\draw (n17) -- +(0, -0.5) -| (n18);
\draw [white, -, line width=6pt] (n17)  +(0, -0.5) -| (n21);
\draw (n17) -- +(0, -0.5) -| (n21);
% \draw (n8) -- +(0, -0.5) -| (n9);
% \draw (n8) -- +(0, -0.5) -| (n12);

\draw [white, -, line width=6pt] (n16) -- +(0, -0.5) -| (n17);
\draw [white, -, line width=6pt] (n16) -- +(0, -0.5) -| (n23);
\draw [white, -, line width=6pt] (n16) -- +(0, -0.5) -| (n25);

\draw (n16) -- +(0, -0.5) -| (n17);
\draw (n16) -- +(0, -0.5) -| (n23);
\draw (n16) -- +(0, -0.5) -| (n25);
% \draw (n2) -- +(0, -0.5) -| (n3);
% \draw (n2) -- +(0, -0.5) -| (n6);
% \draw (n2) -- +(0, -0.5) -| (n8);
% \draw (n1) -- +(0, -0.5) -| (n2);
% \draw (n1) -- +(0, -0.5) -| (n14);
% \draw (n1) -- +(0, -0.5) -| (n16);
% \draw (n0) -- +(0, -0.5) -| (n1);
% \draw (n0) -- +(0, -0.5) -| (n30);
\end{tikzpicture}

%\end{document}
	\caption{A parse tree fragment from Franzen, \emph{The Corrections}.
	Original sentence: something terrible was going to happen.
	}\label{franzenex}
\end{figure}

\section{Task, experimental setup}

We consider a regression problem of a set of novels and their literary ratings.
These ratings have been obtained in a large reader survey
(about 14k participants),\footnote{%
    The survey was part of The Riddle of Literary Quality,
    %The survey was part of the project The Riddle of Literary Quality,
    cf.\ \url{http://literaryquality.huygens.knaw.nl}}
in which 401 recent, bestselling Dutch novels
(as well as works translated into Dutch) where rated on a 7-point Likert scale from
\emph{definitely not} to \emph{highly} literary.
The participants were presented with the author and title of each novel,
and provided ratings for novels they had read.
The ratings may have been influenced by well known authors or titles,
but this does not affect the results of this paper
because the machine learning models are not given such information.
The task we consider is to predict the mean\footnote{%
	Strictly speaking the Likert scale is ordinal and calls for the median,
	but the symmetric 7-point scale
	and the number of ratings
	arguably makes using the mean permissible;
	% w.r.t.\ the median, the mean
	the latter
	provides more granularity and sensitivity to minority ratings.}
rating for each novel.
We exclude 16 novels that have been rated by less than 50 participants.
91~\% of the remaining novels have a $t$-distributed
95~\% confidence interval $<0.5$;
e.g., given a mean of 3, the confidence interval typically ranges
from 2.75 to 3.25. Therefore for our purposes
the ratings form a reliable consensus.
Novels rated as highly literary have smaller confidence intervals,
i.e., show a stronger consensus.
Where a binary distinction is needed, we call a rating of 5 or
higher `literary.'

Since we aim to extract relevant features from the texts themselves
and the number of novels is relatively small, we %need to
apply cross-validation, so as to exploit the data to the
fullest extent while maintaining an out-of-sample approach.
We divide the corpus in 5 folds of roughly equal size,
with the following constraints:
\begin{enumerate*}[label={(\alph*)}]
	\item novels by the same author must be in the same fold,
	    since we want to rule out any influence of author style
	    on feature selection or model validation;
	\item the distribution of literary ratings in each fold should be similar
		to the overall distribution (stratification).
\end{enumerate*}

We control for length and potential particularities of the start of novels by
considering sentences 1000--2000 of each novel.
18 novels with fewer than 2000 sentences are excluded.
Together with the constraint of at least 50 ratings,
this brings the total number of novels we consider to 369.

We evaluate the effectiveness of the features
using a ridge regression model, with 5-fold cross-validation;
we do not tune the regularization.
The results are presented incrementally, to illustrate the contribution of each
feature relative to the features before it. This makes it possible to gauge the
effective contribution of each feature while taking any overlap into account.

We use $R^2$ as the evaluation metric, expressing the percentage of
variance explained (perfect score 100);
this shows the improvement of the predictions over a baseline model that always
predicts the mean value (4.2, in this dataset). A mean baseline model is
therefore defined to have an $R^2$ of 0. Other baseline models, e.g., always
predicting 3.5 or 7, attain negative $R^2$ scores, since they perform worse
than the mean baseline. Similarly, a random baseline will yield
a negative expected $R^2$.

\section{Basic features}

Sentence length, direct speech, vocabulary richness, and compressibility
are simple yet effective stylometric features.
We count direct speech sentences by matching on specific punctuation;
this provides a measure of the amount of dialogue versus narrative text in the novel.
Vocabulary richness is defined as the proportion of words in a text
that appear in the top 3000 most common words of a large reference corpus~(Sonar 500; Oostdijk et al., 2013\nocite{sonar500});
this shows the proportion of difficult or unusual words.
Compressibility is defined as the \texttt{bzip2} compression ratio of the
texts; the intuition is that a repetitive and predictable text will be highly compressible.
\textsc{cliches} is the number of clich\'e expressions in the texts
based on an external dataset of 6641 clich\'es~\cite{wingerden2015cliche};
clich\'es, being marked as informal and unoriginal, are expected to be more
prevalent in non-literary texts.
\autoref{tblbasic} shows the results of these features.
Several other features were also evaluated but were either not effective or
did not achieve appreciable improvements when these basic features are
taken into account; notably
Flesch readability~\cite{flesch1948readability},
average dependency length~\cite{gibson2000dependency},
and D-level~\cite{covington2006complex}.
% Avg.\ dependency length is the mean distance between heads and dependents
% in the syntactic dependencies of the parse trees

\begin{table}[htb]\centering\footnotesize
\begin{tabular}{>{\scshape}p{0.6\linewidth}>{\raggedleft}p{0.2\linewidth}}
                             & $R^2$ \tabularnewline \midrule
mean sent.\ len.\             & 16.4 \tabularnewline
+ \% direct speech sentences  & 23.1 \tabularnewline
+ top 3000 vocab.\            & 23.5 \tabularnewline
+ bzip2\_ratio                & 24.4 \tabularnewline
+ cliches                     & 30.0 \tabularnewline
\end{tabular}
\caption{Basic features, incremental scores.}\label{tblbasic}
\end{table}

\section{Automatically induced features}

In this section we consider extracting syntactic features,
as well as three (sub)lexical baselines.

\textsc{topics} is a set of 50 topic weights induced
with Latent Dirichlet Allocation~(LDA; Blei et al., 2003\nocite{blei2003latent})
from the corpus~(for details, cf.~Jautze et al., 2016\nocite{jautze2016topic}).
% this can be seen as a dimensionality reduction of unigram bag-of-word counts.

Furthermore, we use character and word $n$-gram features.
For words, bigrams present a good trade off in terms of informativeness (a bigram
frequency is more specific than the frequency of an individual word)
and sparsity (three or more consecutive words results in a large number of
$n$-gram types with low frequencies).
For character $n$-grams, $n=4$ achieved good performance
in previous work~(e.g., Stamatatos, 2006\nocite{stamatatos2006charngrams}).

We note three limitations of $n$-grams.
First, the fixed $n$: larger or discontiguous chunks are
not extracted. Combining $n$-grams does not help since a linear model
cannot capture feature interactions, nor is the consecutive occurrence of two
features captured in the bag-of-words representation.
Second, larger $n$ imply a combinatorial explosion of possible features,
which makes it desirable to select the most relevant features.
Finally, word and character $n$-grams are surface features without linguistic abstraction.
One way to overcome these limitations is to turn to syntactic parse trees
and mine them for relevant features unrestricted in size.

Specifically, we consider tree fragments as features, which are arbitrarily-sized fragments
of parse trees. If a parse tree is seen as consisting of a sequence of grammar
productions, a tree fragment is a connected subsequence thereof.
Compared to bag-of-word representations, tree fragments can capture
both syntactic and lexical elements; and these combine to represent
constructions with open slots (e.g., to take NP into account),
or sentence templates (e.g., ``Yes, but \dots'', he said).
Tree fragments are thus a very rich source of features, and larger
or more abstract features may prove to be more linguistically interpretable.

We present a data-driven method for extracting and selecting tree fragments.
Due to combinatorics, there are an exponential number of possible fragments
given a parse tree. For this reason it is not feasible to extract all fragments
and select the relevant ones later; we therefore use a strategy to directly
select fragments for which there is evidence of re-use by considering
commonalities in pairs of trees.
This is done by extracting the largest common syntactic fragments from pairs of
trees~\cite{sangatiefficiently,vancranenburgh2014linear}.
This method is related to tree-kernel methods~\cite{collins2002kernels,moschitti2006treekernel},
with the difference that it extracts an explicit set of fragments.
The feature selection approach is based on relevance and
redundancy~\cite{yu2004efficient}, similar to \newcite{swanson2013extracting}.
\newcite{kim2011treemining} also use tree fragments, for authorship attribution,
but with a frequent tree mining approach; the difference with our approach
is that we extract the largest fragments attested in each tree pair,
which are not necessarily the most frequent.

\subsection{Preprocessing}\label{secpreproc}

We parse the 369 novels with Alpino~\cite{bouma2001alpino}.
The parse trees include discontinuous constituents, non-terminal labels
consist of both syntactic categories and function tags, selected morphological
features,\footnote{The \textsc{dcoi} tag set~\cite{eynde2005pos} is
	fine grained; we restrict the set to distinguish the 7 coarse \textsc{pos}
	tags, as well as infinite verbs, auxiliary verbs, proper nouns,
	subordinating conjunctions, personal pronouns, and postpositions.}
and constituents are binarized head-outward with a markovization of
$h$=1, $v$=1~\cite{klein2003accurate}.

For a fragment to be attested in a pair of parse trees, its labels need to
match exactly, including the aforementioned categories, tags, and features.
The $h=1$ binarization implies that fragments may contain partial constituents;
i.e., a contiguous sequence of children from an $n$-ary constituent.

\autoref{franzenex} shows an example parse tree; for brevity, this tree is
rendered without binarization. The non-terminal labels consist of a syntactic
category (shown in red), followed by a function tag (green).
The part-of-speech tags additionally have morphological features (black) in
square brackets.
Some labels contain percolated morphological features, prefixed by a colon.

\subsection{Mining syntactic tree fragments}

The procedure is divided in two parts.
The first part concerns fragment extraction:
\begin{enumerate}[noitemsep]
	\item
		Given texts divided in folds $F_1 \dots F_n$, each $C_i$ is the
		set of parse trees obtained from parsing all texts in $F_i$.
		Extract the largest common fragments of the parse trees in all pairs
		of folds $\langle C_i, C_j \rangle$ with $i<j$.
		A common fragment $f$ of parse trees $t_1, t_2$ is a connected subgraph
		of $t_1$ and $t_2$.
		The result is a set of initial candidates that occur in at least
		two different texts,
		stored separately for each pair of folds $\langle C_i, C_j \rangle$.
	\item Count occurrences of all fragments in all texts.
\end{enumerate}

Fragment selection is done separately w.r.t.\ each test fold.
Given test fold $i$, we consider the fragments found in training folds
$\{1..n\} \;\setminus\; i$; e.g., given $n=5$, for test fold 1 we select only from the
fragments and their counts as observed in training folds 2--5.
Given a set of fragments from training folds, selection proceeds as follows:
\begin{enumerate}[noitemsep]
	\item Zero count threshold: remove fragments that occur in less
		than 5~\% of texts (too specific to particular novels);
		frequency threshold: remove fragments that occur less than 50 times
		across the corpus (too rare to reliably detect a correlation with the ratings).
		%These thresholds were chosen as reasonable lower bounds
		%for when there is definitely not enough information
		%to select a fragment as relevant.
	\item Relevance threshold: select fragments by considering the
		correlation of their counts
		with the literary ratings of the novels in the training folds.
		Apply a simple linear regression based on the Pearson correlation
		coefficient, and use an F-test to filter out fragments
		whose $p$-value\footnote{If
			we were actually testing hypotheses we would need to apply
			Bonferroni correction to avoid the Family-Wise Error due to
			multiple comparisons; however, since the regression here is
			only a means to an end, we leave the $p$-values uncorrected.}
		$> 0.05$. The F-test determines significance based on the number of
		datapoints $N$, and the correlation $r$; the effective threshold
		is approximately $|r| > 0.11$.
	\item Redundancy removal: greedily select the most relevant fragment and
		remove other fragments that are too similar to it.
		Similarity is measured by computing the correlation coefficient
		between the feature vectors of two fragments,
		with a cutoff of $|r| > 0.5$.
		Experiments where this step was not applied indicated
		that it improves performance.
\end{enumerate}

Note that there is some risk of overfitting since fragments are both
extracted and selected from the training set. However, this is
mitigated by the fact that fragments are extracted from pairs of folds, while
selection is constrained to fragments that are attested and significantly
correlated across the whole training set.

The values for the thresholds were chosen manually and not tuned,
since the limited number of novels is not enough to provide a proper tuning set.
\autoref{numfragsperstep} lists the number of fragments extracted from
folds 2--5 after each of these steps.

\begin{table}[htb]\centering\footnotesize
	\begin{tabular}{p{0.6\linewidth}>{\raggedleft}p{0.2\linewidth}}
		% folds 2-5
		\toprule
		recurring fragments                 & 3,193,952 \tabularnewline
		% after threshold: 
			occurs in $> 5\%$ of texts      &   375,514 \tabularnewline
		% after threshold:
			total freq.\ $> 50$ across corpus &    98,286 \tabularnewline
		relevance: correlated s.t.~$p<0.05$ &    30,044 \tabularnewline
		redundancy: $|r|<0.5$               &     7,642 \tabularnewline
		\bottomrule
	\end{tabular}
	\caption{The number of fragments in folds 2--5 after each filtering step.
		}\label{numfragsperstep}
\end{table}

\subsection{Evaluation}

Due to the large number of induced features, Support Vector Regression (SVR)
is more effective than ridge regression. We therefore train
a linear SVR model with the same cross-validation setup,
and feed its predictions to the ridge regression model (i.e., stacking).
Feature counts are turned into relative frequencies. %$L_2$-normalized.
The model has two hyper-parameters: $C$ determines the regularization, and
$\epsilon$ is a threshold beyond which predictions are considered good enough
during training.
Instead of tuning these parameters we pick fixed values of
$C$=100 and $\epsilon$=0, reducing regularization compared to the default of
$C$=1 and disabling the threshold.

Cf.\ \autoref{tblregfrag} for the scores.
The syntactic fragments perform best, followed by char.\ 4-grams
and word bigrams.
We report scores for each of the 5 folds separately because the
variance between folds is high. However, the differences between
the feature types are relatively consistent.
The variance is not caused by the distribution of ratings,
since the folds were stratified on this. Nor can it be explained by
the agreement in ratings per novel, since the 95~\% confidence intervals
of the individual ratings for each novel were of comparable width
across the folds.
Lastly, author gender, genre, and whether the novel was translated
do not differ markedly across the folds.
It seems most likely that the novels simply differ in how predictable
their ratings are from textual features.

\begin{table}[t]\centering\footnotesize
\setlength{\tabcolsep}{4pt}
\begin{tabular}{lrrrrrr}
{}          &      1 &      2 &      3 &      4 &      5 &    Mean\\\midrule
Word Bigrams&    59.8&    47.0&    58.0&    63.6&    50.7&    55.8\\
Char.~4-grams&   58.6&    50.4&    54.2&    65.0&\bf 56.2&    56.9\\
Fragments   &\bf 61.6&\bf 53.4&\bf 58.7&\bf 65.8&    46.5&\bf 57.2\\
\bottomrule
\end{tabular}
\caption{Regression evaluation.
    $R^2$ scores on the 5 cross-validation folds.}\label{tblregfrag}
\end{table}

\begin{table}[t]\centering\footnotesize
\begin{tabular}{>{\scshape}p{0.6\linewidth}>{\raggedleft}p{0.2\linewidth}}
                             & $R^2$ \tabularnewline \midrule
Basic features (\autoref{tblbasic}) & 30.0 \tabularnewline
+ topics                      & 52.2 \tabularnewline
+ bigrams                     & 59.5 \tabularnewline
+ char.\ 4-grams              & 59.9 \tabularnewline
+ fragments                   & 61.2 \tabularnewline
\end{tabular}
\caption{Automatically induced features; incremental scores.}\label{tblauto}
\end{table}

In order to gauge to what extent these automatically induced features are
complementary, we combine them in a single model together with the basic
features; cf.\ the scores in \autoref{tblauto}.
Both character 4-grams and syntactic fragments still provide a relatively large
improvement over the previous features, taking into account the inherent diminishing
returns of adding more features.

\begin{figure}[htb]\centering
    \includegraphics[width=0.48\textwidth]{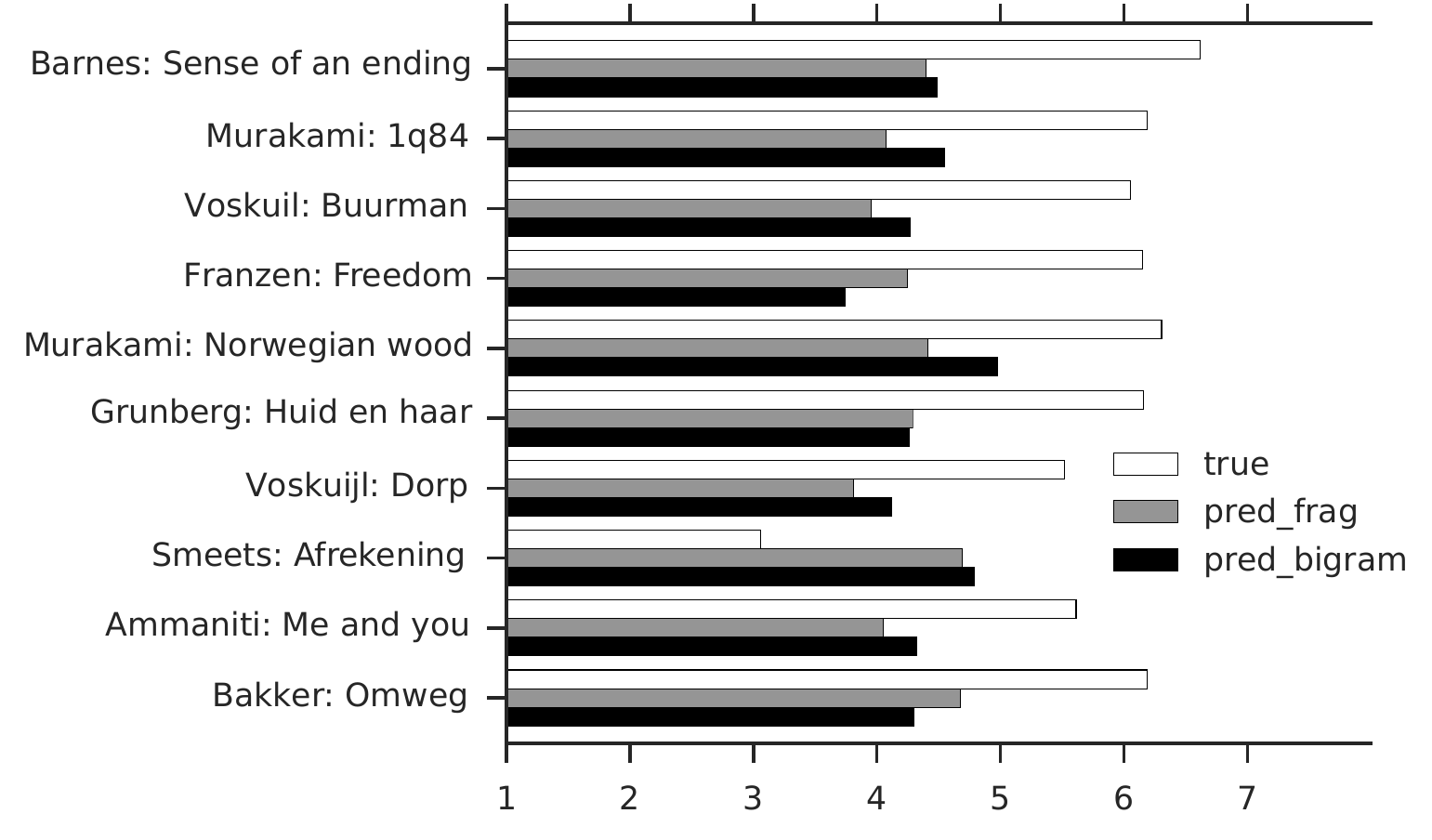}
	\caption{The ten novels with the largest prediction error
		(using both fragments and bigrams).
	}\label{errorbarplot}
\end{figure}

\begin{table}[htb]\centering\scriptsize\tabcolsep=0.11cm
{\setlength{\tabcolsep}{1pt}
\begin{tabular}{%
	l
	>{\raggedleft}b{0.11\linewidth}
	>{\raggedleft}b{0.11\linewidth}
	>{\raggedleft}b{0.11\linewidth}
	>{\raggedleft}b{0.11\linewidth}
	>{\raggedleft}b{0.11\linewidth}
	>{\raggedleft}b{0.11\linewidth}
	}
Novel                          & residual\\(true - pred.) & mean sent.\ len.\ & \% direct speech & \% Top 3000 vocab.\ & bzip2 ratio \tabularnewline \midrule
Rosenboom: Zoete mond          & 0.075 & 23.5 & 24.7 & 0.80 & 0.31 \tabularnewline
Mortier: Godenslaap            & 0.705 & 24.9 & 25.2 & 0.77 & 0.34 \tabularnewline
Lewinsky: Johannistag          & 0.100 & 18.3 & 28.6 & 0.85 & 0.32 \tabularnewline
Eco: The Prague cemetery       & 0.148 & 24.5 & 15.7 & 0.79 & 0.33 \tabularnewline \midrule
Franzen: Freedom               & 2.154 & 16.2 & 56.8 & 0.84 & 0.33 \tabularnewline
Barnes: Sense of an ending     & 2.143 & 14.1 & 23.1 & 0.85 & 0.32 \tabularnewline
Voskuil: Buurman               & 2.117 & 7.66 & 58.0 & 0.89 & 0.28 \tabularnewline
Murakami: 1q84                 & 1.870 & 12.3 & 20.4 & 0.84 & 0.32 \tabularnewline
\end{tabular}}
\caption{Comparison of baseline features for %literary
	novels with good (1--4) and bad (5--8) predictions.}\label{underestimated}
\end{table}

\autoref{errorbarplot} shows a bar plot of the ten novels with the
largest prediction error with the fragment and word bigram models.
Of these novels, 9 are highly literary and underestimated by the model.
For the other novel (Smeets, Afrekening) the literary rating is overestimated
by the model. Since this top 10 is based on the mean prediction from both
models, the error is large for both models. This does not change when the top
10 errors using only fragments or bigrams is inspected; i.e., the hardest
novels to predict are hard with both feature types. 

What could explain these errors?
At first sight, there is no obvious commonality between
the literary novels that are predicted well, or between the ones with a large
error; e.g., whether the novels have been translated or not does not
explain the error.
A possible explanation is that the successfully predicted literary novels share
a particular (e.g., rich) writing style that sets them apart from other novels,
while the literary novels that are underestimated by the model
are not marked by such a writing style.
It is difficult to confirm this directly by inspecting the model,
since each prediction is the sum of several thousand features,
and the contributions of these features form a long tail.
If we define the contribution of a feature as the absolute value of its weight
times its relative frequency in the document, then in case of
Barnes, \emph{The sense of an ending},
the top 100 features contribute only 34~\% of the total prediction.

\begin{figure}[htb]\centering
\includegraphics[width=0.48\textwidth]{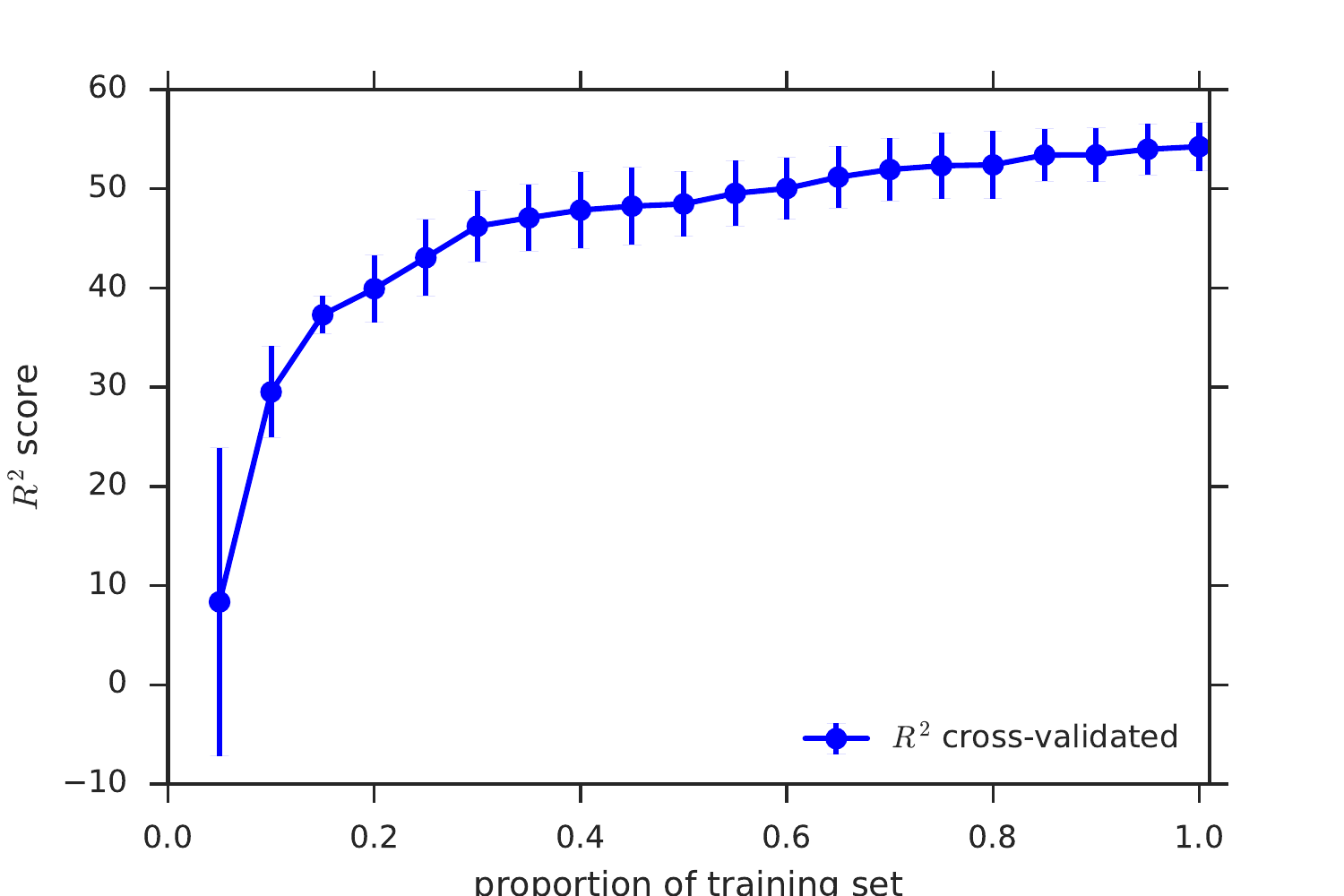}
\caption{Learning curve when varying
	training set size. % (left), features (right).
	The error bars show the standard error.}\label{learningcurves}
\end{figure}

\autoref{underestimated} gives the basic features for the top 4 literary novels
with the largest error and contrasts them with 4 literary novels which are well
predicted.
The most striking difference is sentence length: the underestimated literary
novels have markedly shorter sentences.
Voskuil and Franzen have a higher proportion of direct speech
(they are in fact the only literary novels in the top 10 novels with the most
direct speech). Lastly, the underestimated novels have a higher proportion of
common words (lower vocabulary richness).
These observations are compatible with the explanation
suggested above, that a subset of the literary novels share
a simple, readable writing style with non-literary novels.
Such a style may be more difficult to detect than a literary style with long
and complex sentences, or rich vocabulary and phraseology,
because a simple, well-crafted sentence may not offer overt surface markers
of stylization.
Book reviews appear to support this notion for \emph{The sense of an ending}:
``A slow burn, measured but suspenseful, this compact novel makes every slyly
crafted sentence count''~\cite{tonkin2011sense};
and ``polished phrasings, elegant verbal exactness
and epigrammatic perceptions''~\cite{kemp2011sense}.

\begin{figure*}[htb]\centering
\includegraphics[width=0.7\textwidth]{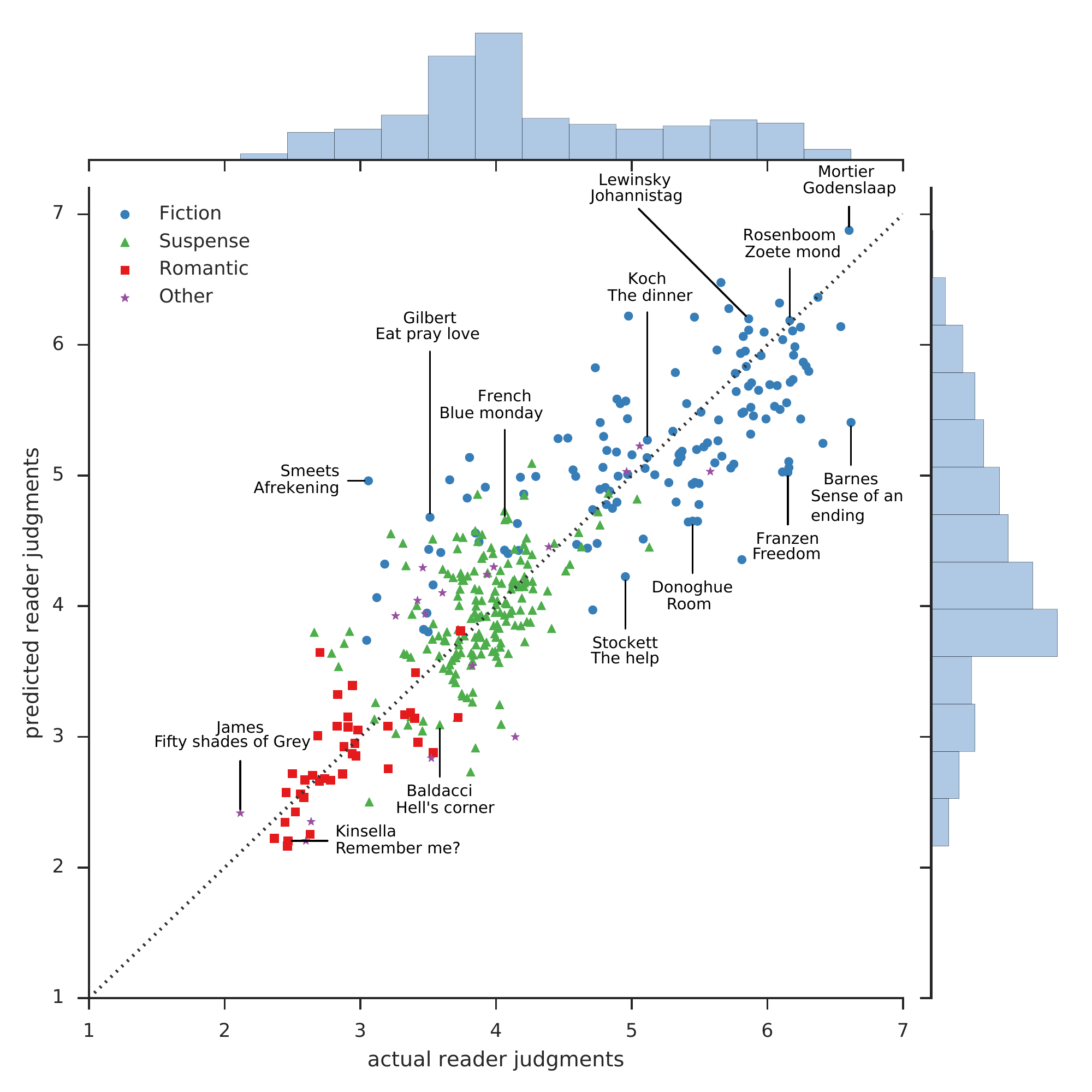}
\caption{A scatter plot of regression predictions and actual
	literary ratings. Original/translated titles.
	Note the histograms beside the axes showing the distribution of ratings (top)
	and predictions (right).
}\label{figregens}
\end{figure*}

In order to test whether the amount of data is sufficient to learn to predict the ratings,
we construct a learning curve for different training set sizes; cf. \autoref{learningcurves}.
The set of novels is shuffled once, so that initial segments of different
size represent random samples.
The novels are sampled in 5~\% increments (i.e., 20 models are trained).
The graphs show the cross-validated scores.

The graphs show that increasing the number of novels has a large effect on
performance. The curve is steep up to 30~\% of the training set, and the
performance keeps improving steadily but more slowly up to the last data
point.
Since the performance is relatively flat starting from 85~\%, we can conclude
that the $k$-fold cross-validation with $k=5$ provides an adequate estimate of
the model's performance if it were trained on the full dataset;
if the model was still gaining performance significantly with more training
data, the cross-validation score would underestimate the true prediction
performance.

A similar experiment was performed varying the number of features.
Here the performance plateaus quickly and reaches
an $R^2$ of 53.0~\% at 40~\%, and grows only slightly from that point.

\section{Metadata features}

In addition to textual features, we also include three (categorical)
metadata features not extracted from the text, but still an inherent
feature of the novel in question: \textsc{Genre}, \textsc{Translated}, and
\textsc{Author gender}; cf.~\autoref{tblmeta} for the results.
\autoref{figregens} shows a visualization of the predictions in a scatter plot.

\textsc{Genre} is the coarse genre classification Fiction, Suspense, Romantic,
Other, derived from the publisher's categorization.  Genre alone is already a
strong predictor, with an $R^2$ of 58.3 on its own.  However, this score is
arguably misleading, because the predictions are very coarse due to the
discrete nature of the feature.

A striking result is that the variables \textsc{Author gender}
and \textsc{Translated} increase the score, but only when
they are both present.
Inspecting the mean ratings shows that translated novels by female authors
have an average rating of 3.8, while originally Dutch male authors are rated 5.0
on average; the ratings of the other combinations lie in between these extremes.
This explains why the combination works better than either feature on its own,
but due to possible biases inherent in the makeup of the corpus,
such as which female or translated authors
are published and selected for the corpus,
no conclusions on the influence
of gender or translation should be drawn from these datapoints.

\begin{table}[t!]\centering\footnotesize
\begin{tabular}{>{\scshape}p{0.6\linewidth}>{\raggedleft}p{0.2\linewidth}}
                             & $R^2$ \tabularnewline \midrule
Basic features (\autoref{tblbasic}) & 30.0 \tabularnewline
+ Auto.\ induced feat.\ (\autoref{tblauto}) & 61.2 \tabularnewline
+ Genre                        & 74.3 \tabularnewline
+ Translated                   & 74.0 \tabularnewline
+ Author gender                & 76.0 \tabularnewline
\end{tabular}
\caption{Metadata features; incremental scores.}\label{tblmeta}
\end{table}

\begin{figure*}\centering
	{\centering
% (NP-obj1|<r:let/LET[]> (N[]-hd 0=) (LET[] 1=,))
\begin{tikzpicture}[scale=0.6, align=center,
				% text width=1.5cm, inner sep=0mm, node distance=1mm]
				minimum height=1.25em, text height=1.25ex, text depth=.25ex,
				inner sep=0mm, node distance=1mm]
\footnotesize\sffamily
\path
	(1, 2) node (n0) { \textcolor{blue}{NP}-\textcolor{darkgreen}{obj1} }
	(-1, 1) node (n00) { \dots }
	(0, 1) node (n1) { \textcolor{blue}{N}-\textcolor{darkgreen}{hd} }
	(0, 0) node (n2) { \textcolor{black}{\dots} }
	(2, 1) node (n3) { \textcolor{blue}{LET} }
	(2, 0) node (n4) { \textcolor{red}{,} }
	(3, 1) node (n5) { \dots }
;
\draw (n3) -- +(0, -0.5) -| (n4);
\draw (n1) -- +(0, -0.5) -| (n2);
\draw (n0) -- +(0, -0.5) -| (n1);
\draw (n0) -- +(0, -0.5) -| (n3);
\draw (n0) -- +(0, -0.5) -| (n00);
\draw (n0) -- +(0, -0.5) -| (n5);
\end{tikzpicture}
\hspace{2em}
% (ROOT (LET[] 0=') (SMAIN (NP-su 1=) (SMAIN 2=)) (LET[] 3=.))
\begin{tikzpicture}[scale=0.6, xscale=1.5, align=center,
				minimum height=1.25em, text height=1.25ex, text depth=.25ex,
				inner sep=0mm, node distance=1mm]
\footnotesize\sffamily
\path
	(1, 3) node (n0) { \textcolor{blue}{ROOT} }
	(0, 1) node (n1) { \textcolor{blue}{LET} }
	(0, 0) node (n2) { \textcolor{red}{'} }
	(2, 2) node (n3) { \textcolor{blue}{SMAIN} }
	(1, 1) node (n4) { \textcolor{blue}{NP}-\textcolor{darkgreen}{su} }
	(1, 0) node (n5) { \textcolor{black}{\dots} }
	(3, 1) node (n6) { \textcolor{blue}{SMAIN} }
	(3, 0) node (n7) { \textcolor{black}{\dots} }
	(4, 1) node (n8) { \textcolor{blue}{LET} }
	(4, 0) node (n9) { \textcolor{red}{.} }
;
\draw (n8) -- +(0, -0.5) -| (n9);
\draw (n6) -- +(0, -0.5) -| (n7);
\draw (n4) -- +(0, -0.5) -| (n5);
\draw (n1) -- +(0, -0.5) -| (n2);
\draw (n3) -- +(0, -0.5) -| (n4);
\draw (n3) -- +(0, -0.5) -| (n6);
\draw (n0) -- +(0, -0.5) -| (n1);
\draw (n0) -- +(0, -0.5) -| (n3);
\draw (n0) -- +(0, -0.5) -| (n8);
\end{tikzpicture}
\hspace{2em}
% (PP-mod (VZ[init]-hd 0=) (NP-obj1 (LID[]-det 1=een) (N[]-hd 2=)))
\begin{tikzpicture}[scale=0.6, xscale=1.5, align=center,
				minimum height=1.25em, text height=1.25ex, text depth=.25ex,
				inner sep=0mm, node distance=1mm]
\footnotesize\sffamily
\path
	(1, 3) node (n0) { \textcolor{blue}{PP}-\textcolor{darkgreen}{mod} }
	(0, 1) node (n1) { \textcolor{blue}{VZ}-\textcolor{darkgreen}{hd} }
	(0, 0) node (n2) { \textcolor{black}{\dots} }
	(2, 2) node (n3) { \textcolor{blue}{NP}-\textcolor{darkgreen}{obj1} }
	(1.4, 1) node (n4) { \textcolor{blue}{LID}-\textcolor{darkgreen}{det} }
	(1.4, 0) node (n5) { \textcolor{red}{een} }
	(3, 1) node (n6) { \textcolor{blue}{N}-\textcolor{darkgreen}{hd} }
	(3, 0) node (n7) { \textcolor{black}{\dots} }
;
\draw (n6) -- +(0, -0.5) -| (n7);
\draw (n4) -- +(0, -0.5) -| (n5);
\draw (n1) -- +(0, -0.5) -| (n2);
\draw (n3) -- +(0, -0.5) -| (n4);
\draw (n3) -- +(0, -0.5) -| (n6);
\draw (n0) -- +(0, -0.5) -| (n1);
\draw (n0) -- +(0, -0.5) -| (n3);
\end{tikzpicture}
}
	\includegraphics[width=0.9\textwidth]{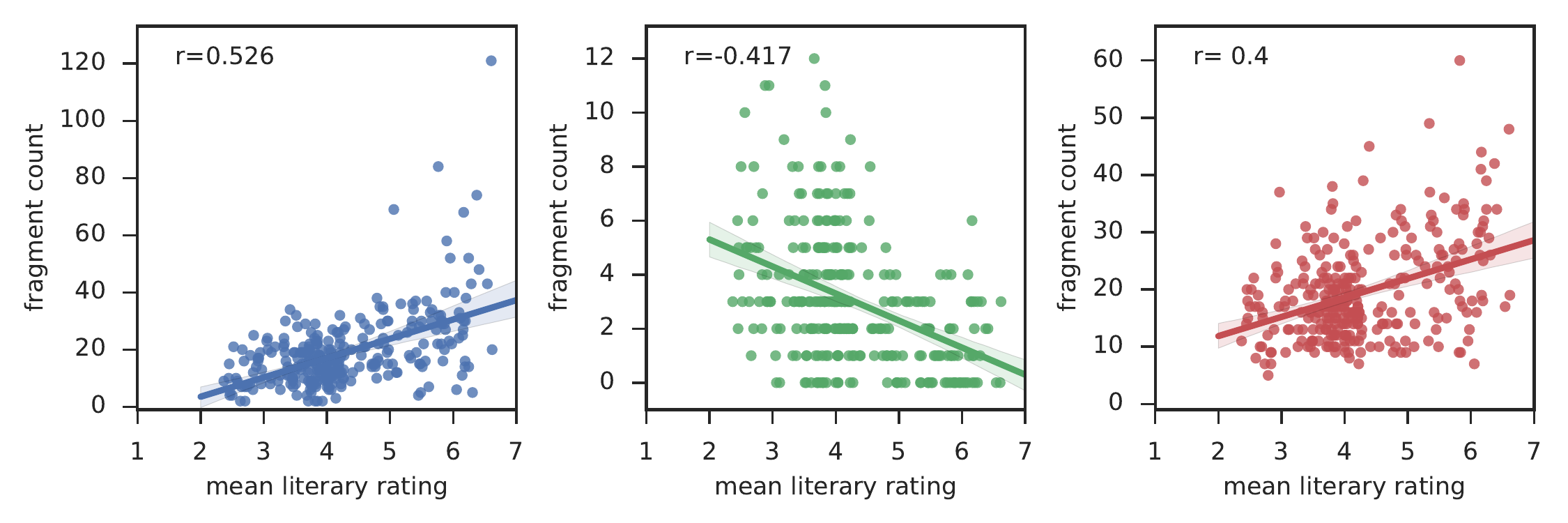}
	\caption{Three fragments whose frequencies in the first fold have a high
		correlation with the literary ratings.
        Note the different scales on the $y$-axis.
		From left to right;
		Blue: complex NP with comma;
		Green: quoted speech;
		Red: Adjunct PP with indefinite article.
		}\label{exfragreg}
\end{figure*}

\section{Previous work} %\section{Related work}

\autoref{tbloverview} shows an overview of previous work on the task of
predicting the (literary) quality of novels. Note that the datasets and targets
differ, therefore none of the results are directly comparable.  For example,
regression is a more difficult task than binary classification, and recognizing
the difference between an average and highly literary novel is more difficult
than distinguishing either from a different domain or genre (e.g., newswire).

\newcite{louwerse2008computationally} discriminate literature from other texts using
Latent Semantic Analysis.
\newcite{ashok2013success} use bigrams, \textsc{pos} tags, and grammar productions to
predict the popularity of Gutenberg texts.
% FIXME Capitalize "van":
\newcite{vancranenburgh2015identifying} predict the literary ratings of texts,
as in the present paper, but only using bigrams, and on a smaller, less diverse corpus.
Compared to previous work, this paper gives a more precise estimate of
how well shades of literariness can be predicted from a diverse range of features,
including larger and more abstract syntactic constructions.

\begin{table}[htb]\centering\footnotesize\tabcolsep=0.11cm
\begin{tabular}{%
	>{\raggedright}p{0.15\textwidth}
	>{\raggedright}p{0.22\textwidth}
	>{\raggedleft}p{ 0.05\textwidth}
	}
Binary classification  & Dataset, task    & Acc.\ \tabularnewline \midrule
\newcite{louwerse2008computationally} & 119 all-time literary classics and 55 other texts,
		literary novels vs.~non-fiction/sci-fi & 87.4 \tabularnewline
\newcite{ashok2013success} & 800 19th century novels, low vs.\ high download count & 75.7 \tabularnewline
\newcite{vancranenburgh2015identifying} & 146 recent novels, low vs.\ high survey~ratings & 90.4 \tabularnewline
\tabularnewline
Regression result & Dataset, task & $R^2$ \tabularnewline \midrule
\newcite{vancranenburgh2015identifying} & 146 recent novels, survey~ratings & 61.3 \tabularnewline
This work & 401 recent novels, survey~ratings & 76.0 \tabularnewline
\end{tabular}
\caption{Overview of previous work on modeling (literary) quality of
	novels.}\label{tbloverview}
\end{table}

\begin{table}[ht]\centering\footnotesize
	\begin{tabular}{p{0.6\linewidth}>{\raggedleft}p{0.2\linewidth}}
		\toprule
		fully lexicalized             & 1,321 \tabularnewline
		syntactic (no lexical items)  & 2,283 \tabularnewline
		mixed                         & 4,038 \tabularnewline
		\midrule
		discontinuous                 &   684 \tabularnewline
		discontinuous substitution site & 396 \tabularnewline
		\midrule
		total                         & 7,642 \tabularnewline
		\bottomrule
	\end{tabular}
	\caption{Breakdown of fragment types selected in the first fold.
		}\label{fragbreakdown}
\end{table}

\section{Analysis of selected tree fragments}

An advantage of parse tree fragments is that they offer
opportunities for interpretation in terms of linguistic aspects
as well as basic distributional aspects such as shape and size.

\autoref{exfragreg} shows three fragments ranked highly by the
correlation metric, as extracted from the first fold.
The first fragment shows an incomplete constituent, indicated by the
ellipses as first and last leaves. Such incomplete fragments are made
possible by the binarization scheme (cf.\ Sec.~\ref{secpreproc}).

\autoref{fragbreakdown} shows a breakdown of fragment types in
the first fold.
In contrast with $n$-grams, we also see a large proportion of purely
syntactic fragments, and fragments mixing both lexical elements and
substitution sites.
In the case of discontinuous fragments, it turns out that the majority
has a positive correlation; this might be due to being associated with
more complex constructions.

\begin{figure}[t!]\centering
\includegraphics[width=0.8\linewidth]{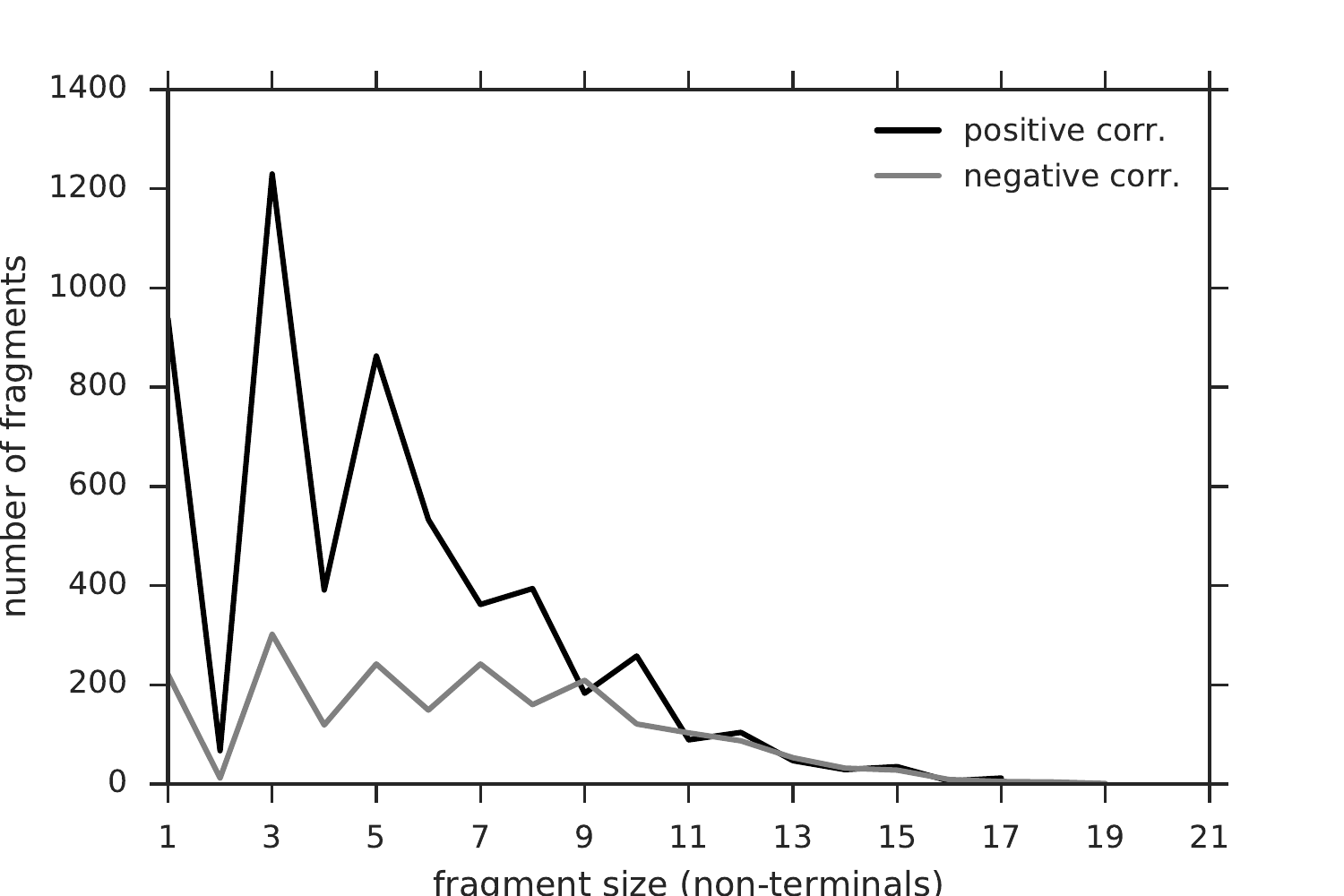}
\caption{Breakdown by fragment size (number of non-terminals).}\label{fragsize}
\end{figure}

\autoref{fragsize} shows a breakdown by fragment size
(defined as number of non-terminals), distinguishing fragments that
are positively versus negatively correlated with the literary ratings.

Note that 1 and 3 are special cases corresponding to
lexical (e.g., DT $\rightarrow$ the)
and binary grammar productions (e.g., NP $\rightarrow$ DT N), respectively.
The fragments with 2, 4, and 6 non-terminals are not
as common because an even number implies the presence of unary nodes.
Except for fragments of size 1, the frontier of fragments can consist of
either substitution sites or terminals (since we distinguish only the number of
non-terminals).
On the one hand smaller fragments corresponding to one or two grammar
productions are most common, and are predominantly positively correlated
with the literary ratings.
On the other hand there
is a significant negative correlation between fragment size and
literary ratings ($r=-0.2, p<0.001$); i.e., smaller
fragments tend to be positively correlated with the literary ratings.

\begin{figure}[htb]\centering
\includegraphics[width=0.8\linewidth]{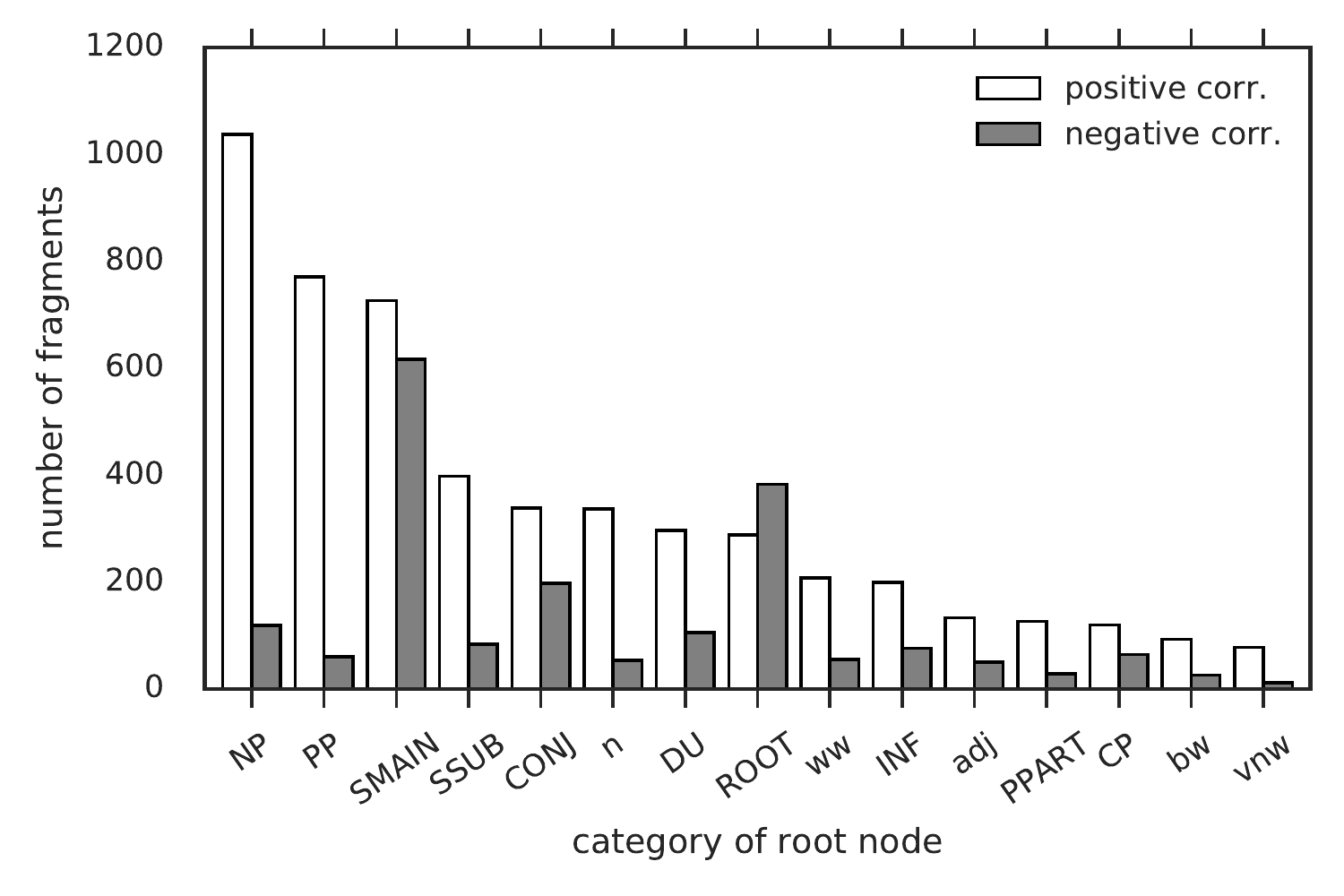} \\
\includegraphics[width=0.8\linewidth]{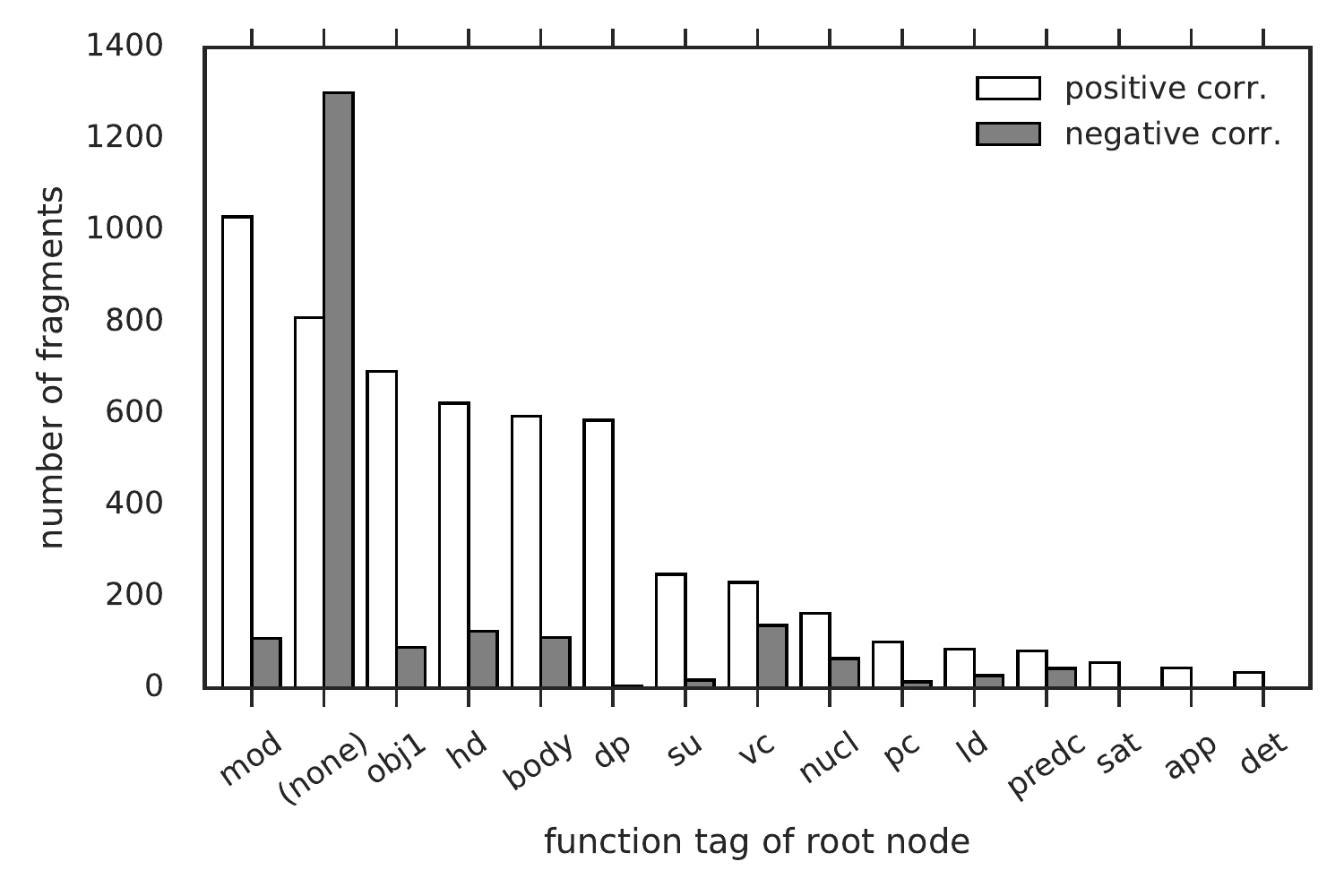}
\caption{Breakdown by category (above) and function tag (below) of fragment
	root (top 15 labels).}\label{fragrootcat}
\end{figure}

It is striking that there are more positively than negatively correlated
fragments, while literary novels are a minority in the corpus
(88 out of 369 novels are rated 5 or higher).
Additionally, the breakdown by size shows that the larger number of positively
correlated fragments is due to a large number of small fragments of size 3 and
5; however, combinatorially, the number of possible fragment types grows
exponentially with size (as reflected in the initial set of recurring
fragments), so larger fragment types would be expected to be more numerous. In
effect, the selected negatively correlated fragments ignore this distribution
by being relatively uniform with respect to size, while the literary fragments
actually show the opposite distribution.

What could explain the peak of positively correlated, small fragments?
In order to investigate the peak of small fragments, we inspect the 40
fragments of size 3 with the highest correlations. These fragments contain
indicators of unusual or more complex sentence structure:
\begin{itemize}[noitemsep]
\item DU, dp: discourse phenomena for which no specific relation
	can be assigned (e.g., discourse relations beyond the sentence level).
\item appositive NPs, e.g., `John the artist.'
\item a complex NP, e.g., containing punctuation, nested NPs, or PPs.
\item an NP containing an adjective used nominally or an infinitive verb.
\end{itemize}
On the other hand, most non-literary fragments are top-level productions
containing ROOT or clause-level labels, for example to introduce direct speech.

Another way of analyzing the selected fragments is by frequency.
When we consider the total frequencies of selected fragments across the corpus,
there is a range of 50 to 107,270.
The bulk of fragments have a low frequency (before fragment selection
2 is by far the most dominant frequency), but the tail is very long.
Except for the fact that there is a larger number of positively correlated
fragments, the histograms have a very similar shape.

Lastly, \autoref{fragrootcat} shows a breakdown by
the syntactic categories and function tags of the root node of the fragments.
The positively correlated fragments are spread over a larger variety of both
syntactic categories and function tags. This means that for most labels, the
number of positively correlated fragments is higher; the exceptions are ROOT,
SV1 (a verb-initial phrase, not part of the top 15), and the absence of a
function tag (indicative of a non-terminal directly under the root node). All
of these exceptions point to a tendency for negatively correlated fragments to
represent templates of complete sentences.

\section{Conclusion}

The answer to the main research question is that literary judgments are
non-arbitrary and can be explained to a large extent from the text itself:
there is an intrinsic \emph{literariness} to literary texts.
Our model employs an ensemble of textual features
that show a cumulative improvement on predictions,
achieving a total score of 76.0~\% variation explained.
This result is remarkably robust:
not just broad genre distinctions, but also finer distinctions in the ratings
are predicted.

The experiments showed one clear pattern:
literary language tends to use a larger set of syntactic constructions
than the language of non-literary novels.
This also provides evidence for the hypothesis
that literature employs a specific inventory of constructions.
All evidence points to a notion of literature
which to a substantial extent can be explained purely
from internal, textual factors,
rather than being %largely
determined by external, social factors.

Code and details of the experimental setup are available at
\url{https://github.com/andreasvc/literariness}

\section*{Acknowledgments}

We are grateful to David Hoover, Patrick Juola, Corina Koolen, Laura Kallmeyer,
and the reviewers for feedback.
This work is part of The Riddle of Literary Quality, a project supported by
the Royal Netherlands Academy of Arts and Sciences through the Computational
Humanities Program.
In addition, part of the work on this paper was funded
by the German Research Foundation DFG \emph{(Deutsche Forschungsgemeinschaft)}.

\phantomsection\addcontentsline{toc}{section}{References}
\makeatletter
\let\@biblabel\@gobble
\makeatother

\bibliographystyle{eacl2017}

\end{document}